\pdfoutput=1

\documentclass[11pt]{article}
\usepackage{amsmath}
\usepackage[final]{EMNLP2023}

\usepackage{times}
\usepackage{latexsym}

\usepackage{graphicx}
\usepackage{enumitem}
\usepackage{multirow}

\usepackage{booktabs}  
\usepackage{multirow}  
\usepackage{siunitx}   

\usepackage[T1]{fontenc}

\usepackage[utf8]{inputenc}

\usepackage{microtype}

\usepackage{inconsolata}

\usepackage[title]{appendix}

\title{SOI Matters: Analyzing Multi-Setting Training Dynamics \\ in Pretrained Language Models via Subsets of Interest}

\author{ \small
\makebox[\textwidth]{%
    \begin{tabular}{c}
        Shayan Vassef\thanks{\hspace{0.4em} Equal Contribution} \\
         University of Illinois Chicago \\
        \texttt{svass@uic.edu}
    \end{tabular} \hspace{2em}
    \begin{tabular}{c}
        Amirhossein Dabiriaghdam\footnotemark[1]
        \\
        University of British Columbia \\
        \texttt{amirhossein@ece.ubc.ca}
    \end{tabular} \hspace{2em}
    \begin{tabular}{c}
        Mohammadreza Bakhtiari\footnotemark[1] \\
        Stony Brook University \\
        \texttt{mohammadreza.bakhtiari@stonybrook.edu}
    \end{tabular}
} \\ \\
\vspace{1em} 
\small
\makebox[\textwidth]{%
    \begin{tabular}{c}
        \textbf{Yadollah Yaghoobzadeh} \\
        \textbf{University of Tehran} \\
        \textbf{\texttt{y.Yaghoobzadeh@ut.ac.ir}}
    \end{tabular}
}
}

\begin{document}
\maketitle
\begin{abstract}

This work investigates the impact of multi-task, multi-lingual, and multi-source learning approaches on the robustness and performance of pretrained language models. To enhance this analysis, we introduce Subsets of Interest (SOI), a novel categorization framework that identifies six distinct learning behavior patterns during training, including forgettable examples, unlearned examples, and always correct examples. Through SOI transition heatmaps and dataset cartography visualization, we analyze how examples shift between these categories when transitioning from single-setting to multi-setting\footnote{For brevity, we use "multi-setting" to refer to multi-task, multi-source, or multi-lingual learning, and "single-setting" to refer to single-task, single-source, or single-lingual learning, respectively.} configurations. We perform comprehensive experiments across three parallel comparisons: multi-task vs. single-task learning using English tasks (entailment, paraphrase, sentiment), multi-source vs. single-source learning using sentiment analysis datasets, and multi-lingual vs. single-lingual learning using intent classification in French, English, and Persian. Our results demonstrate that multi-source learning consistently improves out-of-distribution performance by up to 7\%, while multi-task learning shows mixed results with notable gains in similar task combinations. We further introduce a two-stage fine-tuning approach where the second stage leverages SOI-based subset selection to achieve additional performance improvements. These findings provide new insights into training dynamics and offer practical approaches for optimizing multi-setting language model performance.

\end{abstract}

\section{Introduction}

Deep learning has revolutionized natural language processing (NLP), with Transformer-based models \cite{vaswani2017attention} achieving remarkable success across various tasks. These architectures primarily fall into two categories: decoder-only models, such as GPT-2 \cite{gpt-2}, and encoder-only models, including BERT \cite{bert} and RoBERTa \cite{roberta}. Recently, decoder-only large language models (LLMs) have gained prominence, notably following the success of ChatGPT. While newer open-source LLMs, such as the Llama3 family \cite{dubey2024llama}, facilitate human-friendly interactions across diverse tasks, they do not consistently outperform traditional models \cite{ghaffarzadeh2024large}. Furthermore, current benchmarks for evaluating LLMs emphasize general capabilities such as comprehension and reasoning, frequently neglecting specialized NLP tasks like text classification and named entity recognition. Recent research by \citet{yu2023open} indicates that smaller, fine-tuned encoder-only pretrained language models (PLMs), such as RoBERTa, can match or exceed the performance of larger LLMs across various specialized datasets. Although closed-source LLMs like GPT-4o \cite{hurst2024gpt} can occasionally surpass PLMs with extensive prompt engineering, smaller open-source models offer substantial advantages regarding speed, cost-effectiveness, and transparency. Therefore, systematic analyses of PLM training dynamics remain crucial, even as LLMs increasingly dominate NLP research.

Motivated by this, we systematically investigate the impact of multi-task, multi-lingual, and multi-source learning approaches on the robustness and performance of PLMs. Multi-task learning, which leverages shared knowledge across related tasks, has shown considerable promise for enhancing model generalization and robustness, particularly under constraints of limited data and computational resources. Concurrently, multi-source learning exploits diverse data origins to provide models with a broader understanding of target problems, while multi-lingual learning enables the acquisition of language-agnostic knowledge, significantly improving cross-lingual transfer and performance on low-resource languages.

Despite these advances, a key challenge in training PLMs is handling \textit{Forgettable Examples}, samples that are difficult or out-of-scope, leading models to frequently oscillate between correct and incorrect predictions during training. Although fine-tuning on these challenging examples has proven beneficial in enhancing model robustness \cite{forgettable}, systematic analysis of their underlying learning patterns is currently lacking.

To address this gap, we introduce \textit{Subsets of Interest} (SOI), a novel framework for categorizing dataset samples based on distinct learning behaviors observed during training. Specifically, SOI consists of six categories: \textit{Unlearned Examples} ($\textbf{UNE}$), \textit{Always Correct Examples} ($\textbf{ACE}$), \textit{1-time Forgettable Examples} ($\boldsymbol{1t\text{-}FRGE}$), \textit{At least 2-times Forgettable Examples} ($\boldsymbol{\geq2t\text{-}FRGE}$), \textit{Early-Learned Examples} ($\textbf{ELE}$), and \textit{Late-Learned Examples} ($\textbf{LLE}$). Collectively, these subsets enable detailed insights into the dynamics of model learning behaviors under single- or multi-setting configurations, spanning different tasks, languages, and sources. Furthermore, we investigate the potential of SOI subsets to enhance out-of-distribution performance through second-stage fine-tuning strategies based on various SOI combinations.

The key contributions of this work are as follows: First, we introduce the SOI framework, systematically classifying training samples into distinct learning behavior subsets (Section~\ref{SOI}). Second, we visualize model learning dynamics via dataset cartography and SOI transition heatmaps, offering intuitive insights into sample-level training behaviors (Sections~\ref{SOI_CRP} and \ref{heatmap}). Third, we provide a comprehensive comparative analysis of multi-task, multi-lingual, and multi-source learning methods, evaluating their impacts on both in-distribution (ID) and out-of-distribution (OOD) performances (Subsection~\ref{sec:FF}). Lastly, we extend our OOD evaluations through second-stage fine-tuning on strategically chosen subsets derived from SOI analyses, demonstrating additional performance gains (Subsection~\ref{sec:FF2}).\footnote{Our code is publicly available at \href{https://github.com/vassef/Analyzing-Forgettable-Examples-of-Language-Models-in-Multi-Task-Multi-Lingual-and-Multi-Source-Mod}{this GitHub repository}. It builds upon the implementation provided \href{https://medium.com/@shahrukhx01/multi-task-learning-with-transformers-part-1-multi-prediction-heads-b7001cf014bf}{here}, adapting Hugging Face's transformers library for multi-setting model training.}

\section{Related Work}

In recent years, extensive research has focused on developing multilingual models as well as models capable of performing multiple NLP tasks simultaneously. Multi-task learning leverages shared representations to jointly optimize model performance across various related tasks, enhancing model generalization, robustness, and computational efficiency. Early foundational work by \citet{early-multi-task-in-nlp} introduced multi-task learning concepts to NLP, illustrating that training multiple tasks concurrently could lead to better feature generalization and more robust representations. Subsequent studies have widely adopted transfer learning techniques \citep{howard-ruder-2018-universal}, demonstrating how pretrained language model knowledge can significantly enhance performance on various downstream NLP tasks. Multi-lingual learning, another promising direction, enables models to gain language-agnostic knowledge to understand, generate, and generalize textual information across multiple languages. \citet{xlm-r} introduced XLM-R, a robust cross-lingual PLM trained on diverse multilingual data, significantly improving performance on low-resource languages and facilitating effective cross-lingual transfer.

Understanding model behavior at the individual example level represents another critical aspect in training language models. The phenomenon of example forgetting, instances where models oscillate between correct and incorrect predictions during training, has been thoroughly investigated by \citet{forgettable}. Their work demonstrated that fine-tuning models specifically on these challenging, forgettable examples can significantly enhance model robustness and generalization on task-specific OOD datasets. Complementary to this perspective, \citet{cartography} proposed dataset cartography, a visualization technique characterizing training samples based on prediction confidence and variability metrics. Their method categorizes data into easy-to-learn, hard-to-learn, and ambiguous regions, providing intuitive insights into model behavior throughout training. They conclude that training the model from scratch on the ambiguous region achieves the best ID and OOD performances compared to other scenario cases, including training on hard-to-learn and forgetting examples.

Inspired by these foundational works, our study introduces the Subsets of Interest (SOI) framework, extending beyond previous categorizations with a finer-grained, analytical perspective. Instead of limiting analysis to three regions, SOI systematically classifies training examples into six distinct learning subsets based on their dynamic behaviors during training. Our comprehensive categorization enriches existing analytical tools, offering nuanced insights into model OOD generalization capability across various multi-task, multi-lingual, and multi-source training scenarios.

\section{Experiments Setup}
In this section we introduce three parallel experimental comparisons: multi-task vs. single-task learning, multi-source vs. single-source learning, and multi-lingual vs. single-lingual learning. For each comparison, we conducted similar experiments to evaluate both performance and generalizability. Our experimental framework encompasses various tasks, languages, datasets, and a unified model architecture detailed below.

\subsection{Tasks, Languages and Sources} Our experimental framework spans across multiple dimensions of learning. In the multi-task learning, we utilize three English tasks: entailment (E), paraphrase (P), and sentiment (S). Entailment and paraphrase tasks require binary decisions on semantic relationships between two textual inputs, while sentiment analysis processes single inputs, allowing us to explore combinations of similar tasks (P \& E) versus dissimilar ones (S \& P, S \& E). For the multi-source learning, we focus on sentiment analysis across different data distributions using English datasets, isolating the effects of data source variation from task variation. In our multi-lingual experiments, we conduct intent classification across French (Fr), English (En), and Persian (Fa). This language selection enables us to examine the impact of script and linguistic similarities, as English and French share common features while Persian differs significantly in both script and structure.

\subsection{Datasets}
We employed several benchmark datasets tailored to different learning settings. For each setting, such as multi-task learning, we construct three pairs of datasets, where each pair includes one in-distribution (ID) and one out-of-distribution (OOD)
dataset. Each ID dataset is divided into training, validation, and test splits, whereas the corresponding OOD dataset is treated as a single evaluation set without internal splits. In the following subsections, we detail the specific datasets chosen for each setting.

\subsubsection{Multi-task Learning}
For entailment, SciTail \cite{scitail} serves as the ID dataset, comprising 23,097 training examples. The OOD counterpart is the RTE training set from the GLUE benchmark \cite{wang2018glue}, comprising 2,490 samples. In the paraphrase detection task, we use the Microsoft Research Paraphrase Corpus (MRPC) \cite{mrpc} as the ID dataset, which includes 3,668 training instances. The OOD dataset in this case is a reduced version of the Quora Question Pairs (QQP) training set from the GLUE \cite{wang2018glue}, subsampled to 4,000 examples.

For sentiment classification, we utilize a modified version of the Twitter US Airline Sentiment dataset \cite{twitter_us_airline}, containing 8,078 samples after removing "neutral" labels to enforce binary sentiment polarity. For the corresponding OOD dataset, we adopt a reduced version of the Stanford Sentiment Treebank (SST-2) \cite{socher-etal-2013-recursive} dataset, limited to 4,000 examples to maintain balance across tasks.

\subsubsection{Multi-source Learning}
For multi-source experiments, we use three sentiment analysis datasets as our ID datasets, each containing 50,000 examples sampled from the full dataset with an 80-10-10 train-eval-test split, resulting in 40,000 training instances per dataset. The IMDB movie reviews dataset \cite{rudra2023sentiment} serves as our first source, the Yelp Reviews dataset \cite{hemalatha2019sentiment} comprises business reviews with binary sentiment labels, and Sentiment140 \cite{habib2021twitter} provides sentiment-labeled Twitter content for social media analysis.

As the OOD dataset, we use the Stanford Sentiment Treebank (SST-2) \cite{socher-etal-2013-recursive}, comprising 5,000 examples. Since all sources share the same task, we use the same OOD dataset for all three sources.

\subsubsection{Multi-lingual learning}
For our multilingual experiments, we adopted three intent classification datasets as the ID datasets: Persian subset of MASSIVE \cite{massive_dataset} (11,514 training examples), \textit{Small} subset of CLINC150 \cite{larson-etal-2019-evaluation} for English (7,600 training samples), and LORIA subset of MIAM \cite{colombo-etal-2021-code} for French (8,465 training samples). 

Since the number of intent classes can differ across datasets in the intent classification task, we translated each ID dataset into Burmese, a very low-resource language, and treated these translated versions as the OOD datasets. For translation, we employed the No Language Left Behind machine translation model\footnote{We used  ``\texttt{nllb-200-3.3B}'' model.} \cite{nllb2022}.

\subsection{Architecture} \label{sec:PLMs}
\begin{figure}
    \centering
    \includegraphics[width=0.55\linewidth]{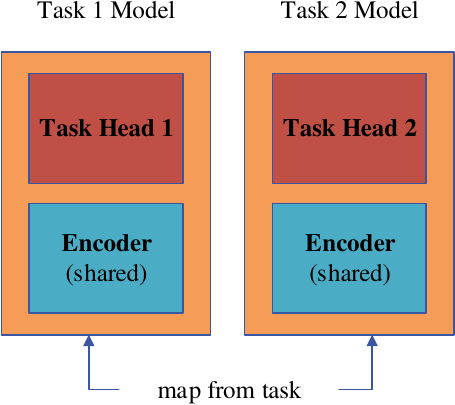}
    \caption{Unified Architecture for Our Multi-Setting Learning Experiments.}
    \label{fig:arc_big_pic}
\end{figure}

Our experiments employ a unified architecture (see Figure~\ref{fig:arc_big_pic}) that leverages shared knowledge through a common encoder. For multi-task and multi-source experiments, we use BERT-base\footnote{We used ``\texttt{bert-base-uncased}'' model.} \cite{bert}, while multi-lingual experiments utilize the multilingual XLM-R\footnote{We used ``\texttt{xlm-roberta-base}'' model.} \cite{xlm-r} model. Each setting maintains specialized classification heads (task-specific, source-specific, or language-specific) attached to the shared encoder.  For multi-setting experiments, training occurs with pairs of tasks/sources/languages: sentiment-entailment (SE), sentiment-paraphrase (SP), and paraphrase-entailment (PE) for tasks; IMDB-Yelp (IY), Sentiment140-Yelp (SY), and IMDB-Sentiment140 (IS) for sources; and finally, English-Persian (En-Fa), French-English (Fr-En), and French-Persian (Fr-Fa) for languages.

\section{Subsets of Interest} \label{SOII}
In this section, we present a comprehensive framework for analyzing deep learning models through the lens of training dynamics. We introduce the concept of \textit{Subsets of Interest} (SOI), a novel categorization system that partitions training examples based on their unique learning patterns observed during the training process. Our analysis unfolds in three complementary parts: first, we formally define the six distinct SOI categories and their characteristics; second, we employ dataset cartography to visualize how these subsets manifest in the confidence-variability space; and third, we introduce transition heatmaps to track how examples migrate between SOI categories under different training configurations. Together, these components provide a systematic approach to understanding and analyzing the complex dynamics of neural network training.

\subsection{SOI Framework and Definitions} \label{SOI}

In this section, we introduce a novel approach to analyzing deep learning models by extracting specific samples from the training set, based on unique learning patterns observed during training. Based on these patterns, the training set spans six distinct subsets, which we call \textit{Subsets of Interest} (SOI): \textit{1. Unlearned Examples} ($\textbf{UNE}$), \textit{2. Always Correct Examples} ($\textbf{ACE}$), \textit{3. 1-time Forgettable Examples} ($\boldsymbol{1t\text{-}FRGE}$), \textit{4. At least 2-times Forgettable Examples} ($\boldsymbol{\geq\!2t\text{-}FRGE}$), \textit{5. Early-Learned Examples} ($\textbf{ELE}$), and \textit{6. Late-Learned Examples} ($\textbf{LLE}$).

$\textbf{UNE}$ refers to samples that show no sign of learning from a certain point onward in the training process. A representative prediction pattern over ten epochs of fine-tuning, assuming the true label is 1, might be $[1, 0, 0, 0, 0, 0, 0, 0, 0, 0]$. $\textbf{ACE}$ denotes samples that the model finds particularly easy to learn, exhibiting consistently correct predictions across all epochs, such as $[1, 1, 1, 1, 1, 1, 1, 1, 1, 1]$.

$\boldsymbol{1t\text{-}FRGE}$ and $\boldsymbol{\geq\!2t\text{-}FRGE}$ represent samples that undergo forgetting events, inspired by \citeposs{FRGE} work on forgetting dynamics during training. In that framework, our UNE is interpreted as a subset of forgettable examples exhibiting an infinite number of forgetting events, denoted by ${\infty t\text{-}FRGE}$. A more recent study by \citet{forgettable} defined forgettable examples as those that experience at least one forgetting event (i.e., ${\geq\!1t\text{-}FRGE}$), or are never learned at all (i.e., UNE).

In our framework, a forgettable example is defined as one that exhibits at least one forgetting and one recollecting event. A \textit{forgetting event} occurs when a previously correct prediction becomes incorrect in a subsequent epoch, while a \textit{recollecting event} is the reverse, an incorrect prediction followed by a correct one. This distinction ensures that FRGE includes dynamic behavior, separating it from UNE, which lacks recollection. For example, a prediction pattern such as $[0, 1, 0, 0, 0, 1, 0, 1, 0, 0]$ contains three forgetting and two recollecting events, and would be categorized as ${\geq\!2t\text{-}FRGE}$.

$\textbf{ELE}$ and $\textbf{LLE}$ refer to samples that initially elude correct classification but eventually reach a point of consistent accuracy. If the first correct prediction occurs on or before epoch 5, the sample is considered ELE; otherwise, it is categorized as LLE, reflecting late-stage learning. For instance, prediction patterns such as $[0, 0, 1, 1, 1, 1, 1, 1, 1, 1]$ and $[0, 0, 0, 0, 0, 0, 0, 0, 1, 1]$ represent ELE and LLE, respectively. All training was conducted over 10 epochs, meaning the ELE/LLE classification is influenced by this hyperparameter. However, the broader notion of early- vs. late-stage learning generalizes across training durations.

\subsection{SOI Visualization via Dataset Cartography} \label{SOI_CRP}
\begin{figure}
    \centering
    \includegraphics[width=\linewidth]{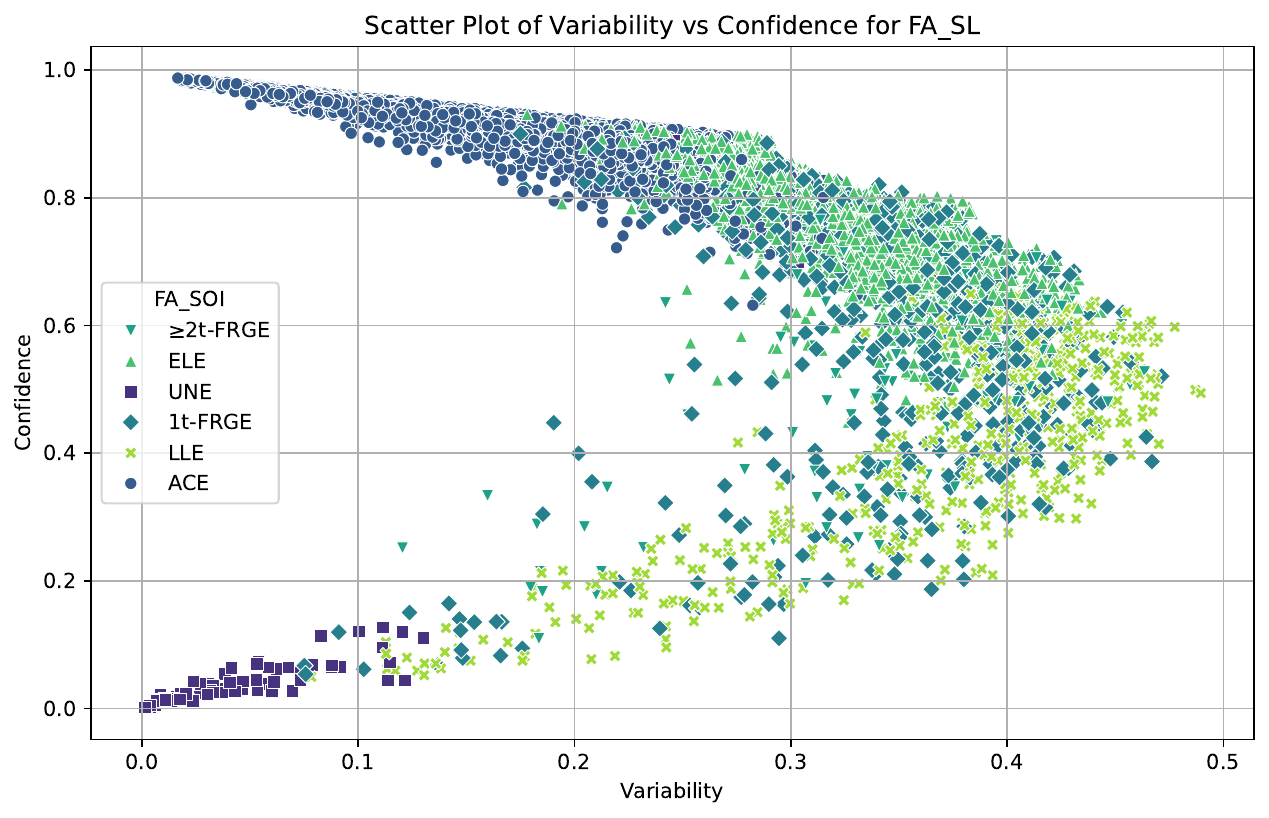}
    \caption{Dataset cartography Map for Single-Lingual Learning in Persian (Fa), showing confidence (average highest prediction probabilities) vs. variability (standard deviation across training epochs).}
    \label{fig:cartography_farsi}
\end{figure}
To better illustrate our definition of SOI, we use dataset cartography analysis, following the approach of \citeposs{cartography}, to visualize how the model learns over time. This method maps training examples onto a two-dimensional space based on two metrics: \textit{confidence} (the average of the model’s highest prediction probabilities) and \textit{variability} (the standard deviation of these predictions across training epochs). This mapping helps us understand how the model behaves with different examples during training.

Cartography divides the examples into three main regions: (1) \textit{easy-to-learn}, with high confidence and low variability; (2) \textit{hard-to-learn}, with low confidence and low variability; and (3) \textit{ambiguous}, with high variability.

Figure~\ref{fig:cartography_farsi} shows the cartography plot for single-lingual learning in Persian. In this plot, the UNE category mainly appears in the hard-to-learn region, while ACE is mostly found in the easy-to-learn region. The LLE class spreads across the hard-to-learn and ambiguous regions, showing a wide range of variability but generally low confidence. On the other hand, ELE stretches from the ambiguous to the easy-to-learn region, suggesting higher confidence even when variability differs. Both ${1t\text{-}FRGE}$ and ${\geq\!2t\text{-}FRGE}$ appear in all three regions, with more examples found in the ambiguous area, which suggests less stable learning behavior. A full version of our cartography visualization is provided in Appendix~\ref{Appendix:cartography}.

\subsection{SOI Transitions through Heatmaps} \label{heatmap}
\begin{figure}[!t]
    \centering    \includegraphics[width=1.15\linewidth]{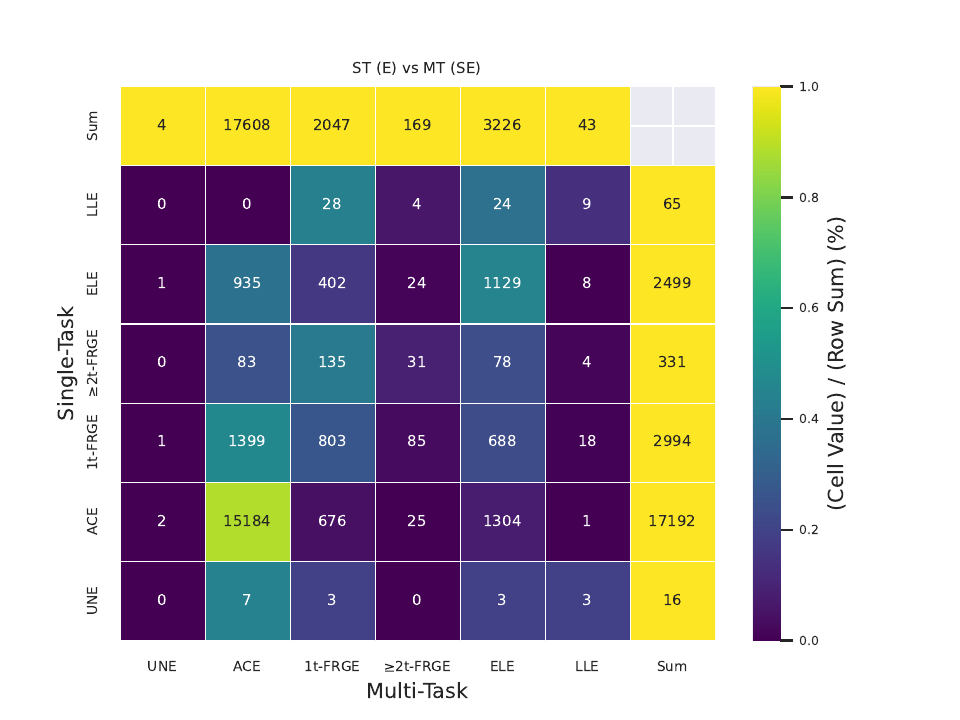}
    \caption{SOI Transition Heatmap: Tracking Training Example Migrations from Single-task (E) to Multi-task (SE). Each cell $H_{i,j}$ shows the number of examples transitioning from SOI category $i$ in single-task to category $j$ in multi-task learning, with the final row and column representing total sums.}
    \label{fig:MT_e_vs_se}
\end{figure}
To analyze how training dynamics evolve under different configurations, we introduce \textit{SOI transition heatmaps}, which capture how examples shift between learning behavior categories when moving from a \textit{single-setting} (e.g., single-task) to a \textit{multi-setting} configuration (e.g., multi-task).

Each transition is represented as a one-to-one mapping from a subset of training examples in a given SOI category under the single-setting to a potentially different category under the multi-setting. This mapping highlights how the training setup influences the model's ability to learn or forget certain examples.

To visualize these transitions, we construct $7 \times 7$ heatmaps. The first 6 rows and columns correspond to the defined SOI categories, while the final row and column represent the \textit{row sums} and \textit{column sums}, respectively. Each cell $H_{i,j}$ in the heatmap records the number of training examples that transitioned from category $i$ (row, under the single-setting) to category $j$ (column, under the multi-setting). For instance in Figure~\ref{fig:MT_e_vs_se}, if 24 examples labeled as LLE in single-task learning become ELE in multi-task learning, then the corresponding cell is $H_{\text{LLE}, \text{ELE}} = 24$.

\section{Results}

In this section, we present our experimental results and analyses, focusing on the impact of different multi-setting configurations on ID and OOD performance, relative to their corresponding single-setting baselines.

\subsection{First-Stage Fine-Tuning} \label{sec:FF}
The goal of first-stage fine-tuning is to adapt PLMs introduced in Subsection~\ref{sec:PLMs} to the in-distribution (ID) training sets under both single-setting and multi-setting configurations. Evaluation is then performed using the corresponding \textbf{ID test sets} and \textbf{OOD test sets}.

\subsubsection{In-Distribution Performance}
Overall, we observed no substantial improvements in \textbf{ID} performance when moving from single-setting to multi-setting fine-tuning. A notable exception was the entailment task, which exhibited a performance gain of 2.6\% when trained jointly with the sentiment task. Additionally, five cases showed marginal improvements (between 0.5\% and 1.0\%), mostly attributed to multi-source learning. The remaining configurations either showed negligible changes ($\leq 0.5\%$) or experienced slight performance degradation (see middle columns of Tables \ref{tab:ML_accuracies}, \ref{tab:MS_accuracies}, and \ref{tab:MT_accuracies}).

\subsubsection{Out-of-Distribution Performance} \label{OOD1}

OOD performance across both single-setting and multi-setting configurations was consistently lower than ID performance. However, the comparative OOD performance between the two configurations revealed insightful trends.

Referencing Table~\ref{tab:ML_accuracies}, we observed similar OOD behavior for French and Persian, while English exhibited a distinct pattern. Specifically, French showed one performance decline (in the French-English pair) and one improvement (French-Persian), while Persian followed a symmetric trend with a drop in the Persian-French case and an increase in Persian-English. In contrast, English experienced performance drops in both of its OOD pairings. These results suggest that multilingual OOD behavior cannot be easily generalized from single-lingual learning. Notably, the positive impact of one language on another’s OOD performance (e.g., Persian improving French) does not imply reciprocal benefit (i.e., French may not enhance Persian).

Turning to Table~\ref{tab:MS_accuracies}, the multi-source learning configuration demonstrated consistent OOD improvements across all six evaluated cases. For Sentiment140, we observed the most significant gain, with a 7\% improvement in OOD accuracy. Other datasets exhibited improvements exceeding 3\%, confirming the effectiveness of multi-source learning in enhancing generalization beyond the training distribution.

Finally, Table~\ref{tab:MT_accuracies} echoes the patterns seen in Table~\ref{tab:ML_accuracies}, with one key distinction: in multi-task learning, when one task enhances another's OOD performance, the improvement is typically mutual. This is evident in the Paraphrase–Entailment configuration (similar tasks), where OOD performance increased by 1.8\% for Entailment and 6.9\% for Paraphrase. In contrast, dissimilar task combinations such as Sentiment–Paraphrase led to performance drops in both tasks under OOD evaluation.

Overall, our experiments reveal that multi-setting fine-tuning, especially in multi-source scenarios, yields clear improvements in OOD performance when datasets originate from the same task. These benefits may also extend to cases where datasets share similar tasks and languages. However, such gains are less predictable when substantial differences exist in linguistic context, highlighting an area for future investigation.

\subsection{Second-Stage Fine-Tuning} \label{sec:FF2}

In the second-stage fine-tuning, we investigate whether the multi-setting models fine-tuned in Subsection~\ref{sec:FF} can be further improved to enhance OOD performance. The fine-tuning sets for this stage are selected based on the heatmaps introduced in Section~\ref{heatmap}, which reveal SOI transitions for a given task, language, or source. For instance, referring to Figure~\ref{fig:MT_e_vs_se}, we can subsample the Entailment training set by extracting all $ELE$ examples from the single-task configuration (e.g., all entries along row $H_{\text{LLE}, \text{-}}$). Using this approach, we experimented with multiple subsampled sets, each defined by a specific heatmap-based criterion, and selected the one that achieved the best average OOD performance across the three multi-task combinations.
 The selected strategy was then applied to the multi-source and multi-lingual setups as well  (see below). Fine-tuning was conducted for 4 epochs, and evaluation was performed solely on OOD sets under multi-setting conditions.

\subsubsection{Heatmap-Based Fine-Tuning Set Selection}\label{op}

We evaluated several fine-tuning set selection strategies based on the transition patterns identified from the heatmaps: \textbf{I.} Transitions representing shifts from more favorable to less favorable learning behaviors (9 out of 36 heatmap transitions: $[ACE, ELE, LLE] \rightarrow 1t\text{-}FRGE$, $[LLE, ELE, ACE, 1t\text{-}FRGE] \rightarrow {\geq\!2t\text{-}FRGE}$, and $[ACE, ELE] \rightarrow LLE$); \textbf{II.} Diagonal entries excluding both $ACE\rightarrow ACE$ and $ELE\rightarrow ELE$; \textbf{III.} Diagonal entries excluding only $ACE\rightarrow ACE$; \textbf{IV.} All forgettable examples identified in single-task learning; \textbf{V.} All forgettable examples identified in multi-task learning; and \textbf{VI.} The entire training set. Among these strategies, method \textbf{III} produced the highest average out-of-distribution (OOD) performance across various multi-task configurations.

\subsubsection{Out-of-Distribution Performance}

We compare the second-stage results against first-stage OOD performance (Subsection~\ref{OOD1}). In multi-lingual learning (Table~\ref{tab:ML_accuracies}), English–French continued to decline, while English–Persian and French–Persian each showed marginal improvements of about 0.3\%. Here, one possible explanation for the overall limited improvements lies in the nature of the OOD test language—\textit{Burmese}, a low-resource language that XLM-R may struggle to represent effectively. As a result, improvements made through training on English, French, or Persian datasets may not transfer well to Burmese, regardless of the fine-tuning strategy. In multi-source learning (Table~\ref{tab:MS_accuracies}), no further gains were observed, likely because the first-stage fine-tuning had already maximized performance. In multi-task learning (Table~\ref{tab:MT_accuracies}), each combination showed a clear improvement for one task and a slight decline for the other. These improvements often occurred where first-stage fine-tuning had previously led to performance drops (e.g., Paraphrase dropped from 62.7\% to 57.3\% in the first stage, then improved to 58.8\%).

Based on our analysis, we found that second-stage fine-tuning was most beneficial in the multi-task setting, had limited or no effect in the multi-source setting, and largely preserved performance in the multi-lingual setting. These results suggest that optimizing the fine-tune set selection with the help of SOI transitions heatmaps is a promising direction for improving OOD robustness in multi-setting configurations.
  
\begin{table*}[!htbp]
\centering
\small
\caption*{Tables~\ref{tab:ML_accuracies}, \ref{tab:MS_accuracies}, and \ref{tab:MT_accuracies} summarize experimental results from initial (first stage) and SOI-guided fine-tuning (second stage). The \textbf{initial fine-tuning} trains PLMs separately (single-setting) or jointly (multi-setting) on in-distribution (ID) datasets; ID columns report in-distribution evaluations, while OOD columns show out-of-distribution performance. The \textbf{SOI-guided fine-tuning} (second stage) further optimizes multi-setting models using targeted subsets strategically selected via SOI transition heatmaps (Subsection~\ref{op}), with improvements measured under the second-stage OOD columns. To interpret these tables, first compare single-setting to multi-setting performances from the initial fine-tuning, then evaluate the additional gains obtained from the subsequent SOI-guided (second stage) fine-tuning.}
\end{table*}

\captionsetup{justification=centering}

\begin{table*}[!htbp]
\centering
\small
\caption{Single and Multi-lingual learning performances. \\ For the multi-lingual setting, we translate each ID dataset (in English, French, or Farsi) into Burmese, and treat the translated samples as OOD evaluation set.}
\label{tab:ML_accuracies}
\begin{tabular}{llccc}
    \toprule
    & & \multicolumn{2}{c}{\textbf{First stage fine-tuning}} & \textbf{Second stage fine-tuning} \\
    \cmidrule(lr){3-4} \cmidrule(lr){5-5}
    \textbf{Model Type} & \textbf{Language} & \textbf{ID} & \textbf{OOD} & \textbf{OOD} \\
    \midrule
    \multirow{3}{*}{Single-lingual} 
    & English & 84.5 & 52.8 & - \\
    & French & 88.5 & 49 & - \\
    & Persian & 87.4 & 62.9 & - \\
    \midrule
    \multirow{2}{*}{Multi-lingual (En-Fr)} 
    & English & 84.4 & 51.9 & 51.8 \\
    & French & 88.7 & 41.6 & 40.9 \\
    \midrule
    \multirow{2}{*}{Multi-lingual (En-Fa)} 
    & English & 84.7 & 48 & 48.1 \\
    & Persian & 87.4 & 63.3 & 63.6 \\
    \midrule
    \multirow{2}{*}{Multi-lingual (Fr-Fa)} 
    & French & 89.4 & 52.2 & 52.2 \\
    & Persian & 87.2 & 61 & 61.4 \\
    \bottomrule
\end{tabular}
\end{table*}

\begin{table*}[!htbp]
\centering
\small
\caption{Single and Multi-source learning performances. \\ OOD dataset: SST-2 is used for all three sources.}
\label{tab:MS_accuracies}
\begin{tabular}{llccc}
    \toprule
    & & \multicolumn{2}{c}{\textbf{First stage fine-tuning}} & \textbf{Second stage fine-tuning} \\
    \cmidrule(lr){3-4} \cmidrule(lr){5-5}
    \textbf{Model Type} & \textbf{Dataset} & \textbf{ID} & \textbf{OOD} & \textbf{OOD} \\
    \midrule
    \multirow{3}{*}{Single-source} 
    & IMDB & 89.4 & 79.4 & - \\
    & Yelp & 93.8 & 79.6 & - \\
    & Sentiment140 & 82.7 & 76 & - \\
    \midrule
    \multirow{2}{*}{Multi-source (IY)} 
    & IMDB & 90.2 & 83.9 & 83.6 \\
    & Yelp & 94.1 & 84.3 & 84.1 \\
    \midrule
    \multirow{2}{*}{Multi-source (SY)} 
    & Sentiment140 & 83.6 & 79 & 79.4 \\
    & Yelp & 93.7 & 83.2 & 82.7 \\
    \midrule
    \multirow{2}{*}{Multi-source (IS)} 
    & IMDB & 90.2 & 85.5 & 84.9 \\
    & Sentiment140 & 83.5 & 83 & 83.1 \\
    \bottomrule
\end{tabular}
\end{table*}

\begin{table*}[!htbp]
\centering
\small
\caption{Single and Multi-task learning performances. \\ OOD dataset: RTE for entailment, QQP for paraphrase, and SST-2 for sentiment.}
\label{tab:MT_accuracies}
\begin{tabular}{llccc}
    \toprule
    & & \multicolumn{2}{c}{\textbf{First stage fine-tuning}} & \textbf{Second stage fine-tuning} \\
    \cmidrule(lr){3-4} \cmidrule(lr){5-5}
    \textbf{Model Type} & \textbf{Task} & \textbf{ID} & \textbf{OOD} & \textbf{OOD} \\
    \midrule
    \multirow{3}{*}{Single-task} 
    & Entailment & 89.3 & 43.9 & - \\
    & Sentiment & 94.6 & 76.7 & - \\
    & Paraphrase & 81.7 & 62.7 & - \\
    \midrule
    \multirow{2}{*}{Multi-task (SP)} 
    & Sentiment & 95 & 75.3 & 74.4 \\
    & Paraphrase & 80.3 & 57.3 & 58.8 \\
    \midrule
    \multirow{2}{*}{Multi-task (SE)} 
    & Sentiment & 95.1 & 62.7 & 64.9 \\
    & Entailment & 91.9 & 38.6 & 38.2 \\
    \midrule
    \multirow{2}{*}{Multi-task (PE)} 
    & Paraphrase & 79.3 & 69.6 & 70 \\
    & Entailment & 89.6 & 45.7 & 45.1 \\
    \bottomrule
\end{tabular}
\end{table*}

\section{Conclusion \& Future Work}
In this work, we conducted a comprehensive investigation into the effects of multi-task, multi-source, and multi-lingual training on PLMs, emphasizing the learning dynamics through the introduction of SOI. By leveraging SOI transition heatmaps and dataset cartography, we provided novel insights into how different training configurations influence both ID and OOD performance. Our results reveal that multi-source learning consistently enhances OOD generalization, while multi-task and multi-lingual learning exhibit more nuanced behavior, offering benefits primarily when task or language similarities exist. The proposed two-stage fine-tuning approach, particularly when guided by SOI-based sample selection, showed further gains in OOD performance, especially in multi-task settings. To sum up, our work highlights the potential of multi-setting configurations in creating more adaptable, robust PLMs capable of generalizing across tasks, languages, and sources.

While our study focused on encoder-based PLMs, future work could apply the SOI framework to large decoder-based language models, such as GPT-style models, to gain insights into their training behaviors and generalization capabilities. Additionally, expanding beyond pairwise combinations to train models on multiple (three or more) tasks, sources, or languages simultaneously could provide a deeper understanding of scaling trends in multi-setting learning. Another direction involves investigating curriculum learning strategies where training is staged according to SOI categories.

\bibliography{custom}
\bibliographystyle{acl_natbib}

\clearpage

\begin{appendices}
\section*{Supplemental Materials}
\section{Experiments Environment} \label{Appendix:eval-setting}
All of the experiments were conducted on the Google Colab virtual systems with around 12.7GB of available RAM and an Nvidia T4 GPU with around 15GB available VRAM.

\section{Complete Dataset Cartography Visualizations} \label{Appendix:cartography}
\begin{figure*}[p]
    \centering
    \includegraphics[width=\textwidth]{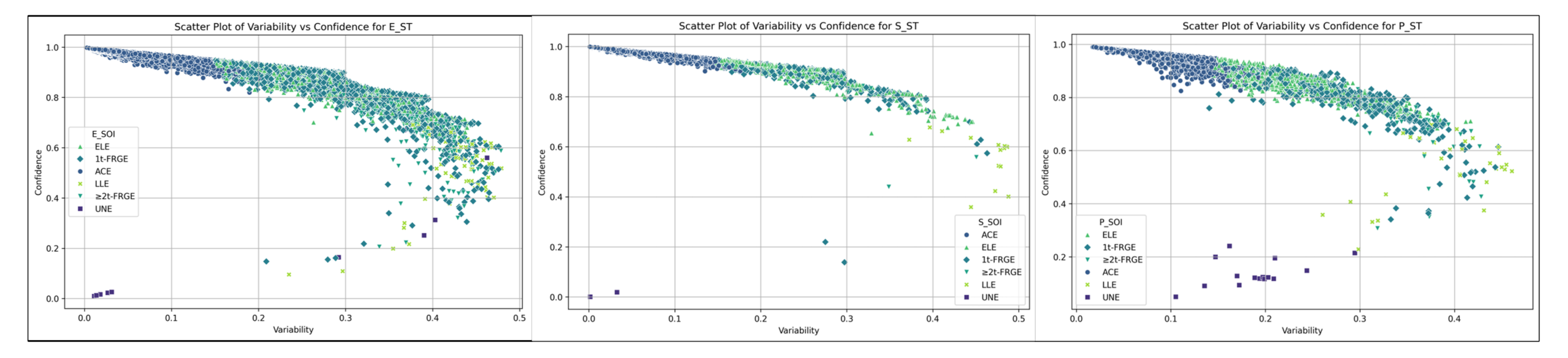}
    \caption{Single-Task Learning (ST) cartography showing the distribution of examples for Entailment, Sentiment, and Paraphrase tasks.}
    \label{fig:stl}
\end{figure*}

\begin{figure*}[p]
    \centering
    \includegraphics[width=\textwidth]{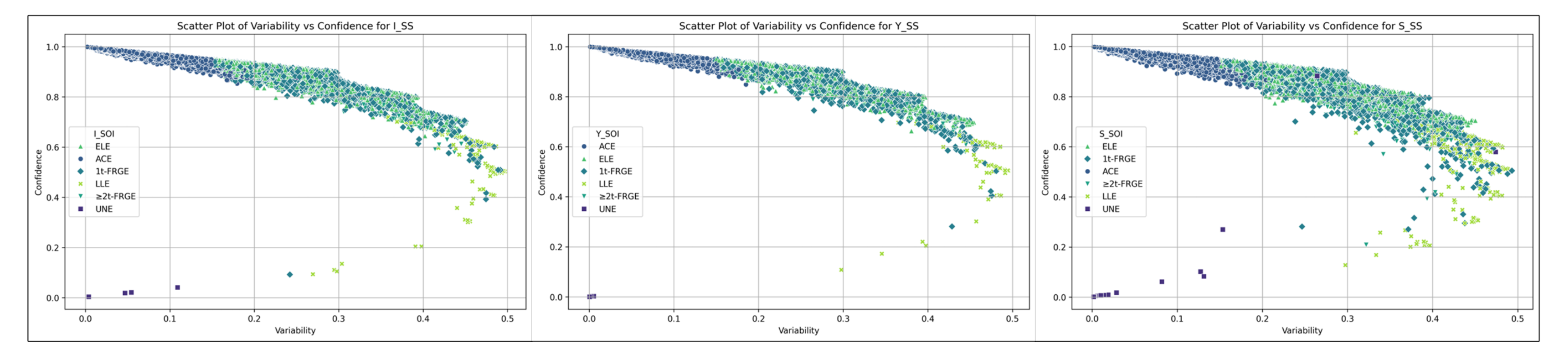}
    \caption{Single-Source Learning (SS) cartography showing the distribution of examples for IMDB, Sentiment140, and Yelp sources.}
    \label{fig:ssl}
\end{figure*}

\begin{figure*}[p]
    \centering
    \includegraphics[width=\textwidth]{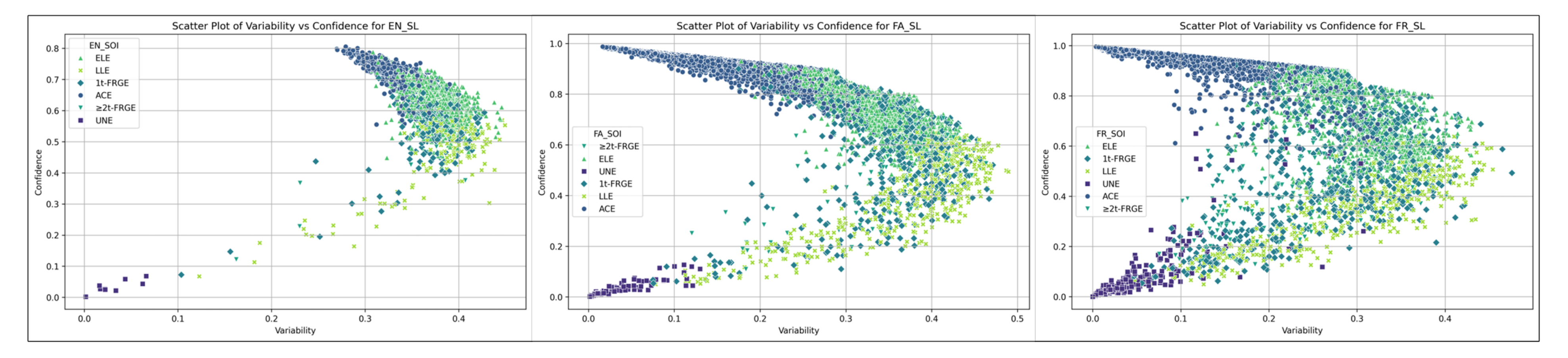}
    \caption{Single-Lingual Learning (SL) cartography showing the distribution of examples for English, French, and Farsi languages.}
    \label{fig:sll}
\end{figure*}

To complement our dataset cartography analysis in Section 5, we provide here the complete set of cartography visualizations across all learning configurations: Figures \ref{fig:stl}, \ref{fig:ssl}, and \ref{fig:sll} present the confidence-variability distributions for single-setting learning across tasks, sources, and languages, respectively. 


\section{Complete Heatmap Visualizations} \label{Appendix:heatmaps}
\begin{figure*}[p]
    \centering
    \includegraphics[width=\textwidth]{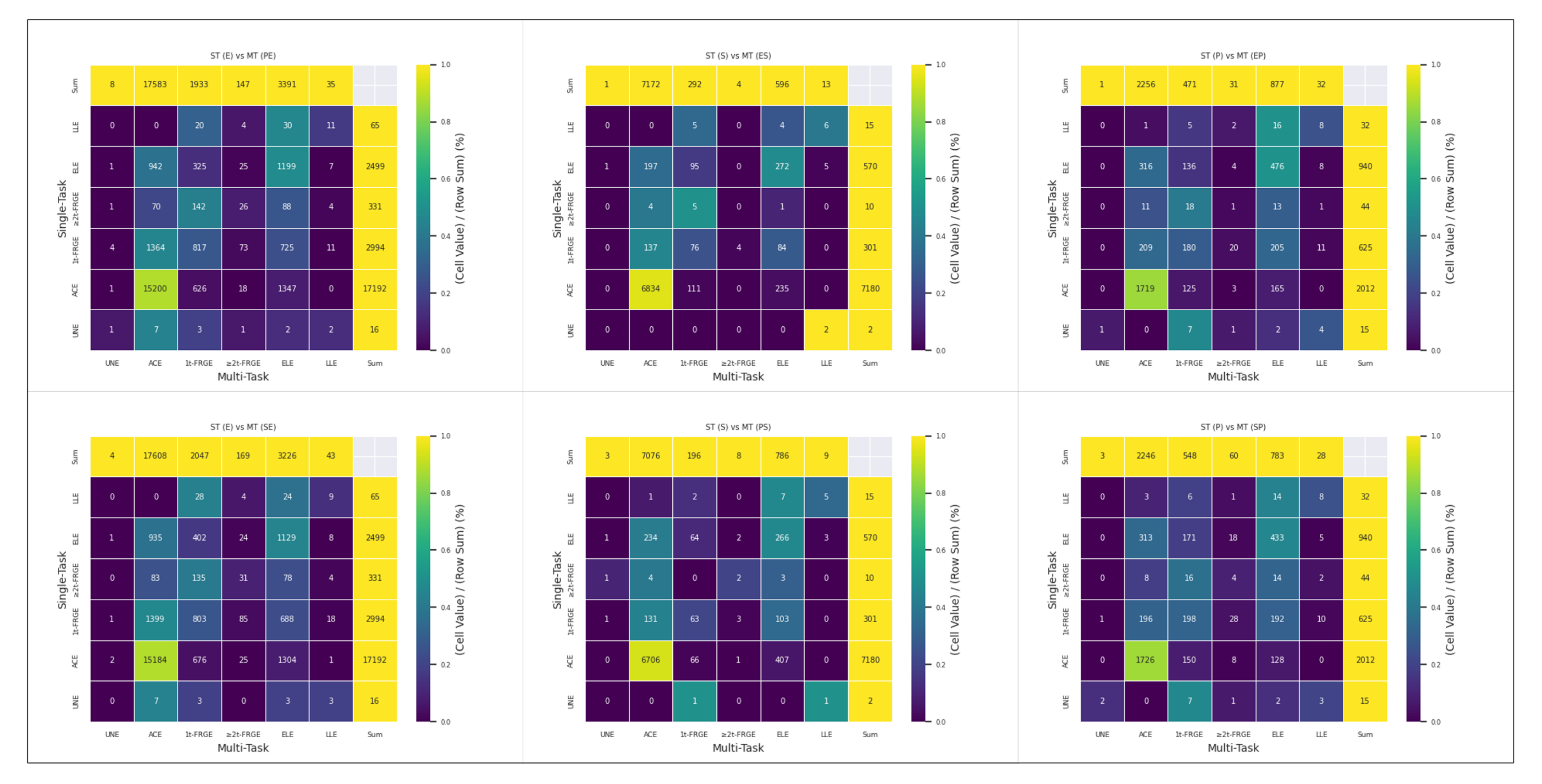}
    \caption{Multi-Task Learning (MT) transition heatmaps showing SOI transitions for all task combinations.}
    \label{fig:mtl_transitions}
\end{figure*}

\begin{figure*}[p]
    \centering
    \includegraphics[width=\textwidth]{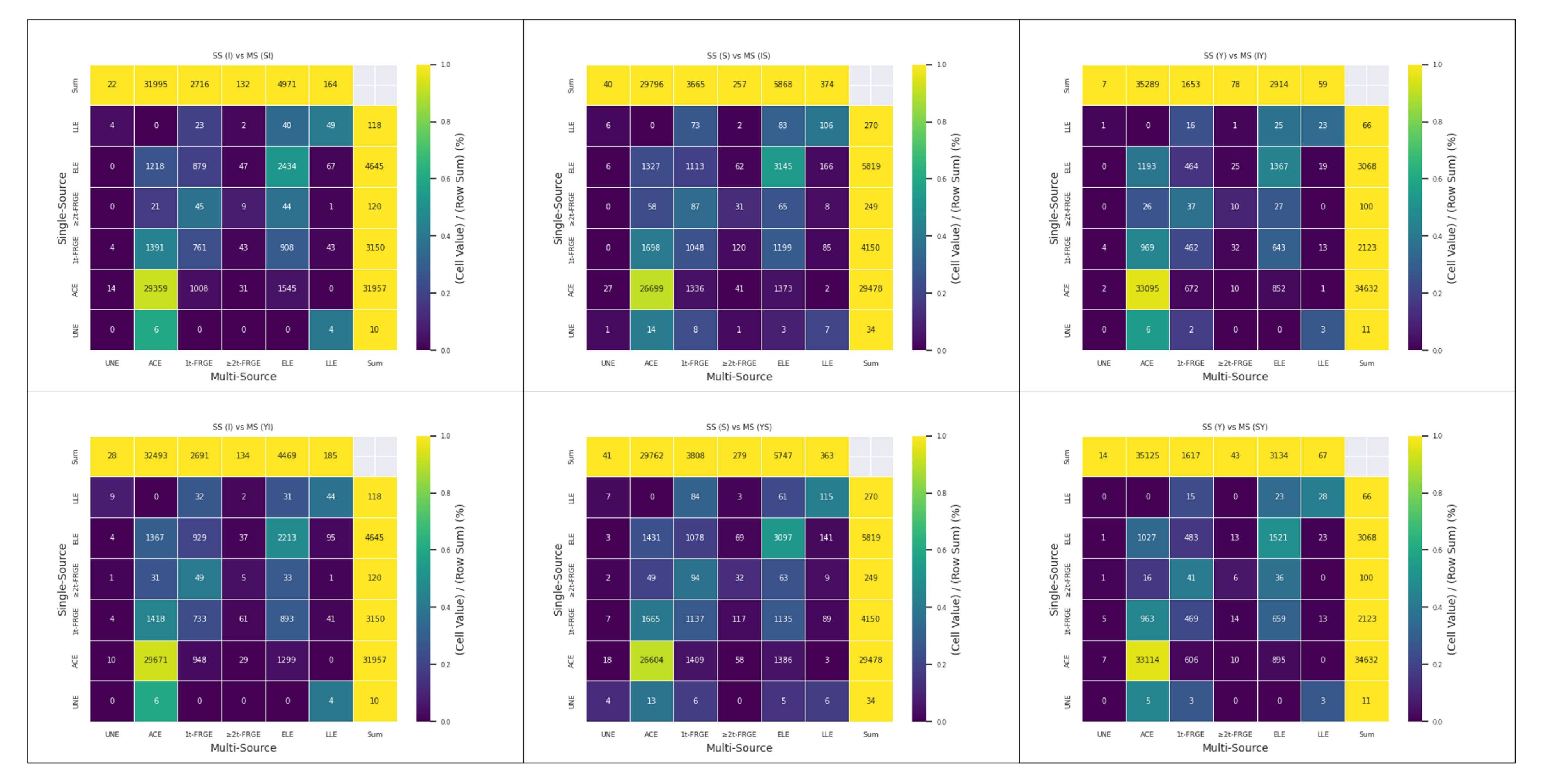}
    \caption{Multi-Source Learning (MS) transition heatmaps showing SOI transitions for all source combinations.}
    \label{fig:msl_transitions}
\end{figure*}

\begin{figure*}[p]
    \centering
    \includegraphics[width=\textwidth]{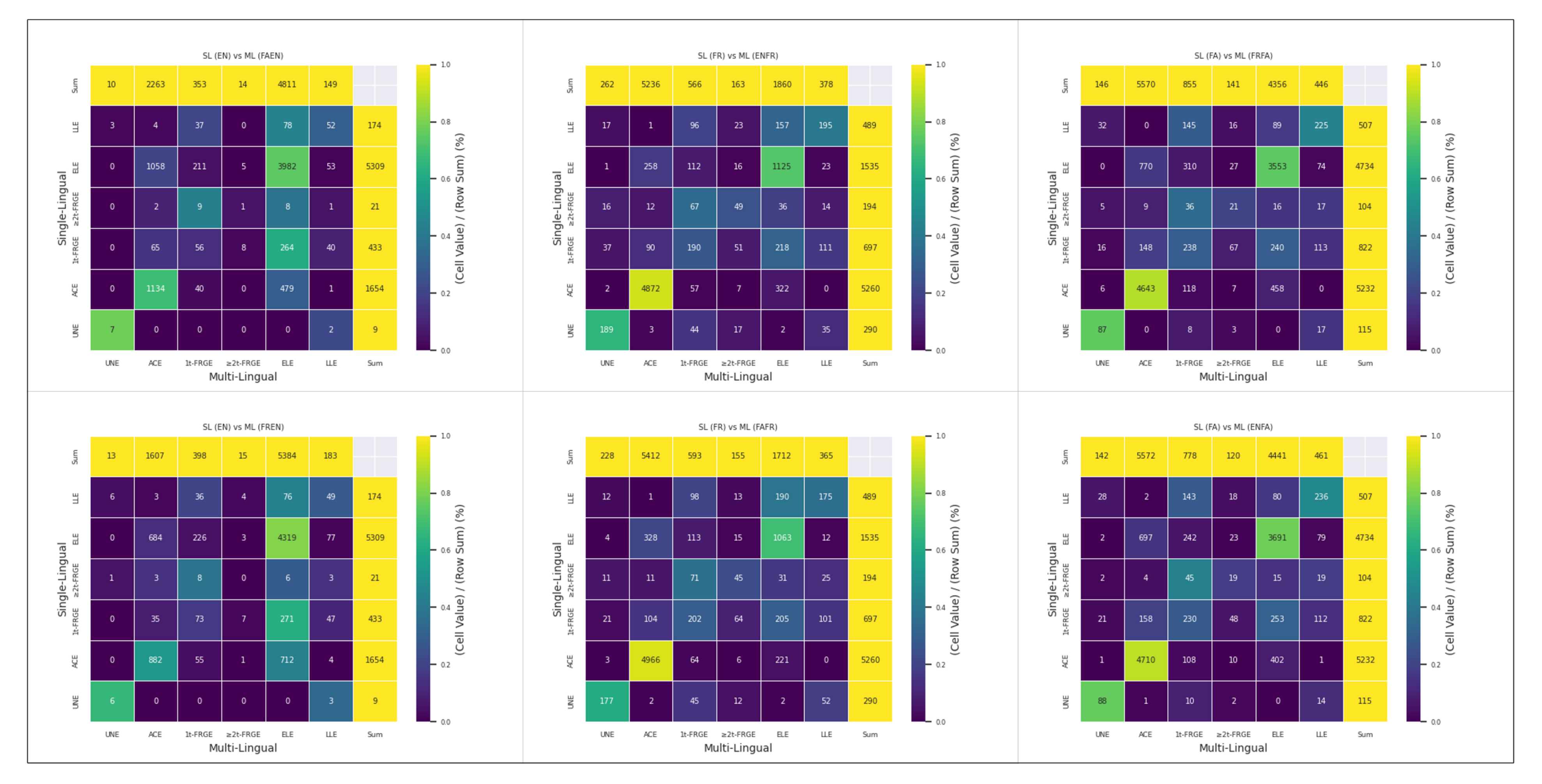}
    \caption{Multi-Lingual Learning (ML) transition heatmaps showing SOI transitions for all language combinations.}
    \label{fig:mll_transitions}
\end{figure*}

As part of our experimental analysis, we generated 18 transition heatmaps - six for each learning mode (multi-task, multi-source, and multi-lingual). While Section 6 presents a detailed analysis of these transitions, here we provide the complete set of heatmaps for reference: Figures \ref{fig:mtl_transitions}, \ref{fig:msl_transitions}, and \ref{fig:mll_transitions} show how samples transition happen between different SOI categories when moving from single-mode to multi-mode learning. Each cell indicates the number of samples that moved from one category to another, with rows representing the initial (single-mode) categories and columns showing the final (multi-mode) categories.
\clearpage
\onecolumn
\twocolumn

\end{appendices}
\end{document}